\documentclass[letterpaper, 10 pt, conference]{ieeeconf}  

\IEEEoverridecommandlockouts                              

\usepackage{amsmath,amssymb,amsfonts}
\usepackage{algorithmic}
\usepackage{graphicx}
\usepackage{braket}
\usepackage{textcomp}
\usepackage{xcolor}
\usepackage{subcaption}
\usepackage{float}
\usepackage{multicol}
\usepackage{caption}
\usepackage{multirow, boldline, bbm, makecell}
\usepackage{hyperref}
\usepackage{colortbl,xcolor}
\usepackage[colorinlistoftodos]{todonotes}

\usepackage[normalem]{ulem} 

\newcommand{\Ztext}[1]{\textcolor{black}{#1}}
\newcommand{\Ctext}[1]{\textcolor{black}{#1}}

\definecolor{MyDarkBlue}{rgb}{0,0.08,1}
\definecolor{MyAqua}{rgb}{0,0.7,0.7}
\definecolor{MyDarkGreen}{rgb}{0.02,0.6,0.02}
\definecolor{MyDarkRed}{rgb}{0.8,0.02,0.02}
\definecolor{MyDarkOrange}{rgb}{0.40,0.2,0.02}
\definecolor{MyPurple}{RGB}{111,0,255}
\definecolor{MyPink}{RGB}{255, 148, 251}
\definecolor{MyRed}{rgb}{1.0,0.0,0.0}
\definecolor{MyGold}{rgb}{0.75,0.6,0.12}
\definecolor{MyDarkgray}{rgb}{0.66, 0.66, 0.66}
\definecolor{nicegreen}{rgb}{0.1, 0.6, 0.2}

\newcommand{\etal}{et al.~}
\newcommand{\theSkin}{RobotSweater}

\title{\LARGE \bf
\theSkin: Scalable, Generalizable, and Customizable Machine-Knitted Tactile Skins for Robots
}

\author{Zilin Si$^{*,1}$, Tianhong Catherine Yu$^{*,1}$, Katrene Morozov$^{2}$, James McCann$^{1}$ and Wenzhen Yuan$^{1}$
\thanks{* Authors with equal contribution.}
\thanks{$^{1}$Zilin Si, Tianhong Catherine Yu, James McCann and Wenzhen Yuan are with Carnegie Mellon University, 5000 Forbes Ave, Pittsburgh, PA, 15213, USA
        {\tt\small <zsi, tianhony, jmccann, wenzheny>@andrew.cmu.edu}}%
\thanks{$^{2}$Katrene Morozov is with University of California, Santa Barbara, Santa Barbara, CA, 93106, USA
        {\tt\small k\_morozov@umail.ucsb.edu}}%
\thanks{This work was supported by CMU Manufacturing Futures Institute and National Science Foundation grant 1955444.}%
}

\let\oldtwocolumn\twocolumn
\renewcommand\twocolumn[1][]{%
    \oldtwocolumn[{#1}{
    \begin{center}
        \includegraphics[width=\linewidth ]{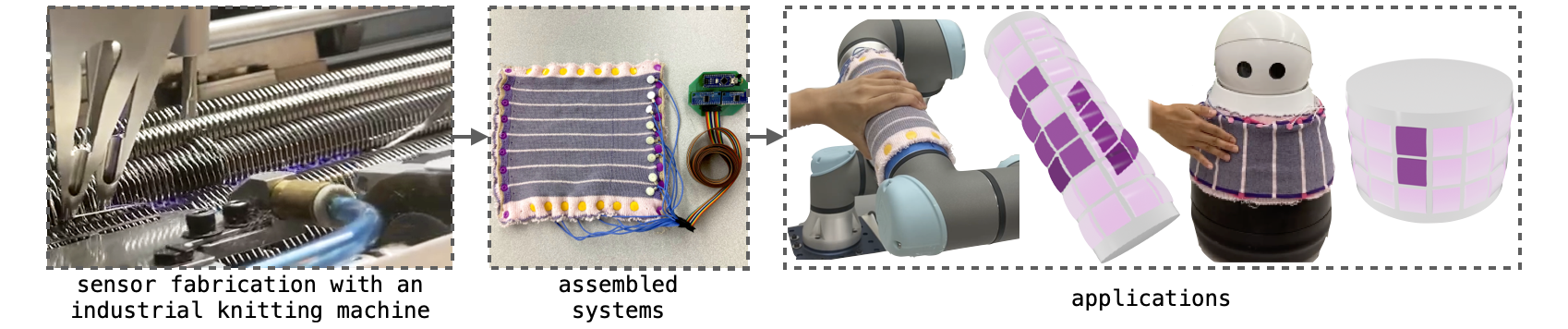}
        \captionof{figure}{\theSkin~is a customizable machine-knitted multi-layer tactile skin. 
        Left: sensor fabrication using an industrial knitting machine. Middle: an assembled tactile skin. 
        Right: applications of our tactile skins including robot arm \Ztext{human lead-through control} and mobile robot human-robot interaction.}
        \label{fig:teaser}
        \vspace{-2mm}
    \end{center}
    }]
}
\begin{document}

\maketitle

\thispagestyle{empty}
\pagestyle{empty}

\begin{abstract}
Tactile sensing is essential for robots to perceive and react to the environment. However, it remains a challenge to make large-scale and flexible tactile skins on robots. Industrial machine knitting provides solutions to manufacture customizable fabrics. 
Along with functional yarns, it can produce highly customizable circuits that can be made into tactile skins for robots. 
In this work, we present \theSkin, a machine-knitted pressure-sensitive tactile skin that can be easily applied on robots. We design and fabricate a parameterized multi-layer tactile skin using off-the-shelf yarns, and characterize our sensor on both a flat testbed and a curved surface to show its robust contact detection, multi-contact localization, and pressure sensing capabilities. The sensor is fabricated using a well-established textile manufacturing process with a programmable industrial knitting machine, which makes it highly customizable and low-cost. The textile nature of the sensor also makes it easily fit curved surfaces of different robots and have a friendly appearance. 
Using our tactile skins, we conduct closed-loop control with tactile feedback for two applications: (1) human lead-through control of a robot arm, and (2) human-robot interaction with a mobile robot.
\end{abstract}

\begin{figure*}[t]
    \centering
    \includegraphics[width=\linewidth]{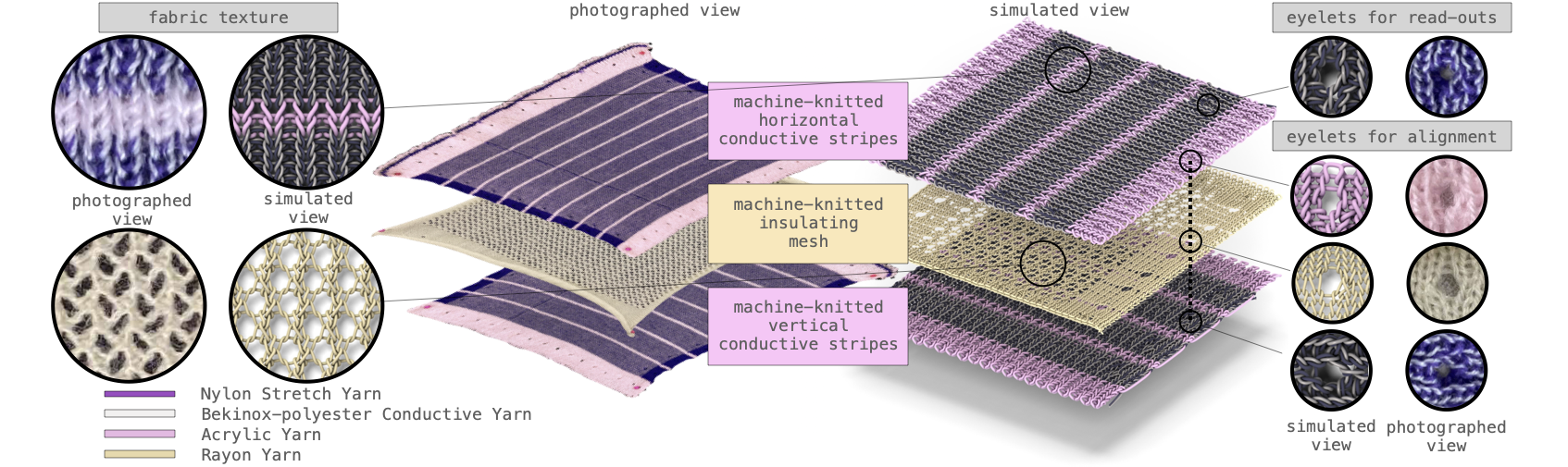}
    \vspace*{-5mm}\caption{An exploded view of the three-layer design: a perforated insulating mesh layer is sandwiched between two orthogonal layers of conductive stripes.
    The conductive stripes form a resistor matrix whose values change in response to applied pressure.
    Eyelets are placed along the borders for alignment and read-out purposes.
    }
    \label{fig:sensor-design}
    \vspace{-5mm}
\end{figure*}

%

\section{Introduction}
Living organisms have skin receptors that help them gather environmental information to react accordingly.
Similarly, robots benefit from skin-like tactile sensors to sense pressure, strain, shear force, temperature, proximity, textures, humidity, etc.~\cite{robot-skin-review} 
For robots, fingertip tactile sensors that sense pressure over a small area have been well developed, yet sensing contact information over a large area on the robot surface remains a challenge. 
The main obstacles are scalability and flexibility. 
Rigid sensors can partially cover a large area using a modular approach to form a multi-modal skin network~\cite{rigid-skin-cell}, but instrumentation and communication are difficult to scale and generalize to different robots.

Flexible and deformable robot skins~\cite{supercapacitive,scalable-stretchable-transistor-array} have desired mechanical properties to adapt to various robots and cover curved surfaces.
One prevalent flexible and deformable material is textile, and the textile manufacturing industry has automated fabrication procedures that generalize to different shapes and produce fabrics at scale.
Previous researchers have made successful industrial machine-knitted tactile sensors~\cite{knitui,knit-human-environment,knit-pipeline-tactile}. 
\Ctext{However, those sensors suffer from non-robust and non-repeatable measurements caused by unmodeled internal deformation, which the contact detection relies on.}
\Ztext{In this work, we propose a novel sensor structure with more robust and repeatable }\Ctext{behaviors}.

We present \theSkin: a\
machine-knitted
pressure-sensitive 
low-cost 
tactile skin that is 
scalable,
generalizable,
and customizable.
The parameterizable design makes the sensor's shape, dimension, and spatial resolution customizable. We fabricate the sensor using off-the-shelf conductive and non-conductive yarns on an automated knitting machine, which makes the cost of the sensor low. The material cost of our sensor is less than \$40, including both electronic readout circuit components ($\sim$\$35) and yarns ($\sim$\$3).
Unlike prior matrix-like resistive tactile sensors~\cite{tapis-magique, scalable-sensor-textile-array}, which sandwich a piezo-resistive layer between two layers of orthogonal highly conductive traces, our sensor design sandwiches an insulating mesh layer between two layers of orthogonal piezo-resistive traces.
This design enables taxel units to be pressure-sensitive under contact and disconnected when there is no contact.
This ensures accurate contact detection and localization which is critial for applications like safe human-robot interaction. The main contributions of this paper are:
\begin{itemize}
    \item We design and manufacture a parameterized machine-knitted resistive-matrix-style tactile sensor and associated read-out circuitry (Section~\ref{sec:fabrication});
    \item we characterize the sensor's properties on both flat and curved surfaces (Section~\ref{sec:characterization});
    \item and we demonstrate the application of the sensors on a robot arm with \Ztext{human lead-through control}, and a mobile robot with human-robot interaction (Section~\ref{sec:applications}).
\end{itemize}

We believe deploying large-scale robot skins can be as easy as putting ``sweaters'' on different robots.
Furthermore, textile-based skins provide unique comfort and familiarity which can promote safer human-robot interaction.

\section{Related Work}
\subsection{Robot Tactile Skins}
Tactile sensors allow robots to measure and react to their surrounding environment. Functions of robot skins include shape-changing, locomotion enabling~\cite{morphing-locomotion-skin}, self-healing~\cite{healing-skin}, and tunable stiffness~\cite{tunable-stiff-skin}. Pressure sensing is an essential way to let the robot feel the world.
Various functional properties have been explored for pressure sensing: piezoresistive, piezocapacitive, triboelectric, iontronic, magnetic, biomimetic, fiber optic, and etc.~\cite{robot-skin-review}.
A challenge in adopting these innovative technologies is the lack of scalable manufacturing methods that are compatible with different robot shapes~\cite{scalable-stretchable-transistor-array}. 
Rigid tactile sensors can cover large areas using modular approaches~\cite{rigid-skin-cell}.
Flexible tactile sensors can further fit on robots with complex geometries~\cite{supercapacitive,stretch-EIT}.

\indent To achieve conformal contacts for surfaces with large curvatures, Ye \etal\cite{conformal-tactile-skin} used porous polyurethane sponge and carbon black to assemble a 4x4 sensing unit for collaborative robots with high sensitivity and fast response time.
Prior works on large-area flexible e-skin commonly use a printing technique~\cite{print-flexible-review} with flexible printed circuits and nanomaterial synthesis~\cite{large-soft-eskin}.
One fabrication technique to make scalable stretchable skin is knitting, and in this work, we use an industrial knitting machine and off-the-shelf functional yarns to manufacture low-cost and scalable tactile skins.

\subsection{Knitted Sensors}
Knitted fabric is intrinsically soft, flexible, conformal, and comfortable.
Knitting with functional yarns yields augmented electronic textiles (e-textiles) with sensing~\cite{sensorKnit} and actuating~\cite{omnifiber} capabilities while preserving the above qualities.
Recent efforts in computational design tools~\cite{knittemplate} and scheduling algorithms~\cite{autoknit} allow researchers to easily program industrial knitting machines and manufacture complex shapes~\cite{knit-compiler}.
Conductive yarns and loop structures of knitted fabric enable strain~\cite{strain-knit-respiratory}, pressure~\cite{knit-pipeline-tactile}, and proximity~\cite{knit-proximity} measurement.
These can be driven by resistive~\cite{knitui}, capacitive~\cite{sensorKnit}, inductive~\cite{knit-inductive}, and impedance~\cite{spray-knit-eit} sensing.
We chose a resistive approach to avoid the capacitive noise in surroundings and the calibration of baseline capacitances~\cite{capacitive-adaptive}.
A matrix readout is chosen to reduce the number of wires, especially in the interior of the sensor ~\cite{knit-humanoid}.
Even though electrical impedance tomography (EIT)~\cite{stretch-EIT} requires a smaller number of electrodes and simpler wiring,
our preliminary tests showed that the tactile sensitivity~\cite{DNN-EIT} and spatial resolution of EIT designs for textile sensors are low.
\\
\indent The most similar approaches to \theSkin~are the following machine-knitted matrix-style pressure-sensitive textiles.
Luo \etal\cite{knit-human-environment} used the inlay knitting technique to integrate novel coaxial piezoresistive fibre into garments to collect spatially dense tactile readings for human-environment interaction learning.
Wicaksono \etal\cite{3DKnITS} machine-knitted conductive traces sandwiching a piezoresistive nylon sheet to fabricate activity recognition socks and mats.
We extend the matrix-style tactile skin idea to simplify the setup process for robot sensing.
We conduct detailed precision and repeatability characterization, which are missing in the previous work. We show the effectiveness of our sensors through closed-loop control systems by using tactile readings on robots. 

\section{Sensor Design and Fabrication}\label{sec:fabrication}
Motivated by the requirement of accurate contact localization and pressure measurement (Sec~\ref{sec:working-principles}), we use a three-layer-knitted-fabric structure for our tactile sensor (Sec~\ref{sec:knitting}): an insulating mesh layer sandwiched between two layers with orthogonal conductive stripes (Fig~\ref{fig:sensor-design}). 
We fabricate our tactile skins with off-the-shelf conductive yarns on a knitting machine, and connect (Sec~\ref{sec:connectors}) the functional textiles to a customized read-out circuit (Sec~\ref{sec:circuit}) to read the resistances of taxel unit at the intersections of two stripes.

\subsection{Working Principles}\label{sec:working-principles}
\begin{figure}[t]
    \centering
    \includegraphics[width=\linewidth]{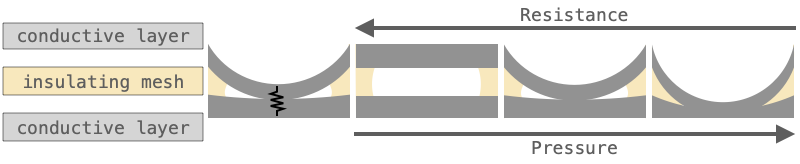}
    \vspace*{-6mm}
    \caption{Each intersection between the horizontal and vertical stripes can be modeled as a variable resistor. The resistance decreases as pressure increases.}
    \label{fig:principle}
\end{figure}

As shown in Fig.~\ref{fig:principle}, the resistances between two layers are decided by the inter-fabric and intra-fabric structural change caused by the layers' deformation.
Without pressure, the top and bottom layers are separated by the insulating layer. 
When a pressure is applied, the top and bottom layers come into contact and the circuit is closed but with high-resistance.
Both the inter-fabric and intra-fabric resistance decreases as the contact pressure increases \Ztext{with an approximate inverse correlation}. The former is caused by the increase in contact areas between different layers, and the later is caused by the changes in loop structure formed by yarns within layers.


\subsection{Machine-knitted Sensing Textiles}\label{sec:knitting}
There are three machine-knitted layers in our sensor design (Fig.~\ref{fig:sensor-design}).
The top and bottom \textit{conductive-stripe layers} have alternating wide conductive and thin insulating stripes.
We keep the insulating stripes narrow to maximize the sensing area and minimize the dead-zone area.
A middle \textit{insulating mesh} layer is sandwiched in between.
We place \textit{eyelets} along the borders of each layer for alignment and circuit connection purposes.
The parameterized knitting patterns are designed and generated in the knitout~\cite{mccann:2017:knitout} language with JavaScript programs and visualized with a web-based tool\footnote{https://textiles-lab.github.io/knitout-live-visualizer/}~\cite{knitout-visualizer}. They are translated to machine instructions and knitted by the machine. 
The knitting layers are fabricated on an industrial v-bed knitting machine, the Shima Seiki SWG091N2, 15 gauge.
\Ctext{The assembled sensor is about 5mm thick.}

\textit{Conductive-Stripe Layers.}
We choose the 1x1 rib structure for high extension potential, good elastic recovery, and minimal strain-induced hysteresis~\cite{rib-strain}.
The insulating stripes separate neighboring conductive stripes.
Insulating stripes are knitted with acrylic yarn (Tamm Petit 2/30).
Conductive stripes are co-knitted with nylon stretch yarn (Maxi Lock Stretch Textured Nylon) and Baekert BK 9036129, a Bekinox-polyester blend in 50/2Nm with a specified resistance of 20 $\Omega$/cm.
The high-elasticity nylon stretch yarn decreases hysteresis~\cite{rib-strain}.
The conductive yarn is plated outside of the nylon yarn (top left in Fig.~\ref{fig:sensor-design}) to maximize conductive contacts and sensitivity when pressure is applied.
The conductive stripes have a smaller stitch and leading setting (25 and 15) than the insulating stripes (35 and 20) due to the different yarn properties.

\textit{Insulating Mesh Layer.}
We use a lace structure, 132 Confetti, as shown in Fig.~\ref{fig:sensor-design} (bottom left) to create the insulating mesh, and a stockinette stitch for denser borders to distinguish the eyelets from the lace holes and provide structural integrity. 
We use rayon yarn (Winning) to knit the middle layer. 
The diameter of the yarn and the size of the mesh holes affect the sensor's sensitivity. 
Sensors with thinner yarn and bigger holes are more sensitive to light touches, while sensors with thicker yarn and smaller holes are more robust to large pressure.
Therefore the mesh layer can be customized to get the desired performance for different purposes.

\begin{figure}[t]
    \centering
    \vspace*{-3.5mm}
    \includegraphics[width=\linewidth]{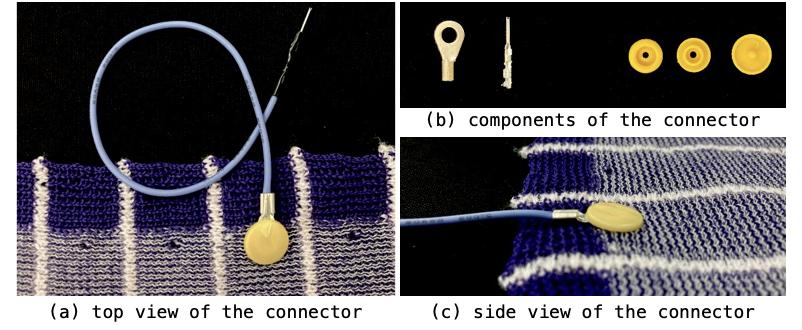}
    \vspace*{-6mm}
    \caption{\Ctext{(a)}: an assembled connector; \Ctext{(b)}: ring terminal, male pin, and snap buttons that are used to connect the fabric to the sensor readout circuit board; \Ctext{(c)}: a side view of the assembled connector on fabric.}
    \label{fig:connector}
\end{figure}
\textit{Eyelets} (Fig.~\ref{fig:sensor-design} right), small holes in fabrics, are machine knitted by transferring loops from one needle to the neighboring one.
Aligning eyelets on different fabric layers helps with consistent layer alignment.
Eyelets on conductive stripes
also serve as connection points to wires for read-outs.

\subsection{Connectors}\label{sec:connectors}
\begin{figure}[t!]
    \centering
    \vspace*{-3mm}
    \includegraphics[width=\linewidth]{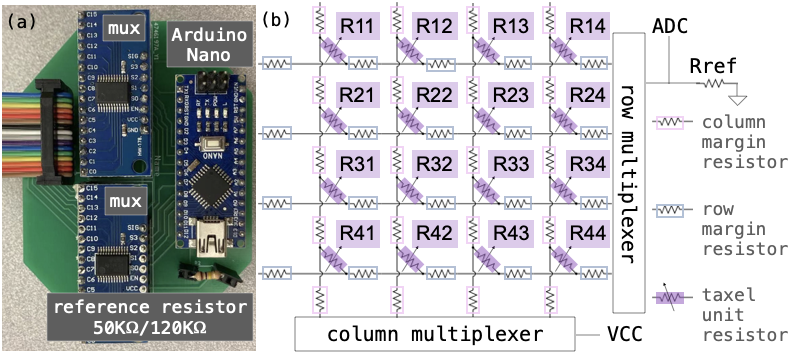}
    \vspace*{-4mm}
    \caption{\Ctext{(a)}: the read-out board contains an Arduino nano, two multiplexers and a reference resistor; and it reads values of the resistor matrix.
    \Ctext{(b)}: the schematic of the readout circuit.
    Due to the highly resistive yarn, row/column margin resistors of resistance \Ctext{$\sim3K\Omega$} are not negligible.
    Behaviors of taxel unit variable resistors are explained in Fig.~\ref{fig:principle}.}
    \label{fig:schematic}
    \vspace*{-5mm}
\end{figure}

Efficiently and reliably connecting hard electrical components to textile-based sensors is an open research problem~\cite{connectors-review}.
We choose snap fasteners, which are commonly used to fasten clothes and are good candidates for e-textile connectors~\cite{wire-popper}.
We stick the snap fastener cap through the read-out hole and ring terminal, then connect to the socket/stud (Fig.~\ref{fig:connector}).
This forms robust and stable contacts between the conductive fabric and ring terminals.
Furthermore, the fasteners are used to button the sensors onto robots. 

\subsection{Read-out circuit}\label{sec:circuit}

We design the circuit to read out the resistance values (Fig.~\ref{fig:schematic}).
It contains an Arduino Nano board, two 16-channel multiplexers (CD74HC4067), and a reference resistor (50K$\Omega$ for the 4x4 prototype and 120K$\Omega$ for other prototypes in this paper). 
Controlled by the row multiplexer, VCC (5V) is switched to connect to different \Ctext{column} stripes.
Controlled by the \Ctext{row} multiplexer, ground (GND) is switched to connect to different \Ctext{row} stripes.
\Ctext{This allows reading $N^2$ taxel units using $2N$ connectors.}
Between the sensor and GND, we add a reference resistor as a voltage divider.
\Ctext{The reference resistor is connected in series with a resistor network consisting of column/row margin resistors and taxel unit resistors}.
For example in Fig.~\ref{fig:schematic}, switching to the 3rd row and the 2nd column reads taxel unit resistor, R32, along with 2 column- and 3 row-margin resistors.
We use the Arduino's internal analog-to-digital converter (ADC) to read \Ztext{tactile readings directly from raw \texttt{analogRead()} outputs,} which measures the voltage difference across the reference resistor.
\Ztext{An increase in the tactile reading ($tr$) corresponds to a decrease resistor network resistance}\Ctext{: $R_{network}=R_{ref}\times (\frac{1024}{tr}-1)$, where 1024 is the maximum ADC output}.


\section{Sensor Characterization}\label{sec:characterization}
\begin{table}[b!]
\centering
\small
\vspace*{-4mm}
\caption{Characteristics of three \theSkin~ prototypes. Sizes are measured in the unstretched condition.}
\begin{tabular}{l|l l l} 
\Xhline{1pt}
 \theSkin~dimension & 4x4 & 3x16 & 8x8 \\
 \hline
 taxel unit size (mm) & 22x22 & 42x42 & 18x18 \\ 
 \theSkin~size (mm) & 145x145 & 150x900 & 180x180 \\ 
 insulating margin (mm) & 3 & 5 & 3\\
 Arduino microcontroller& Uno  & Nano & Nano\\ 
 sample rate (Hz) & 150 & 57 & 42 \\ 
 energy consumption (mW) & 125.75 & 57.7 & 56.85 \\ 
\Xhline{1pt}
\end{tabular}
\label{table:sensor-specs}
\end{table}
Sensor characterization is essential to understand the effectiveness of the sensor and provides guidance for applications. 
We fabricated three different \theSkin~prototypes for characterization and applications, as summarized in Table~\ref{table:sensor-specs}.
We characterize our tactile skin sensors on both a flat testbed and a curved testbed, as shown in Fig.~\ref{fig:flat-ft}. We evaluate our sensors based on their sensitivity,
repeatability, 
contact localization precision, 
robustness under different contact shapes,
and readout circuit performance.

For all of the characterization experiments, we use an indenter \Ctext{to apply normal forces} to our tactile sensors. We attach a 6-axis force-torque sensor, either an ATI Nano 17 or a Nordbo NRS-6050-D50 sensor at the back of the indenter to measure the ground truth force.

\subsection{Force sensitivity}\label{sec:force-sensitivity}
\begin{figure}[t!]
    \centering
    \includegraphics[width=\linewidth]{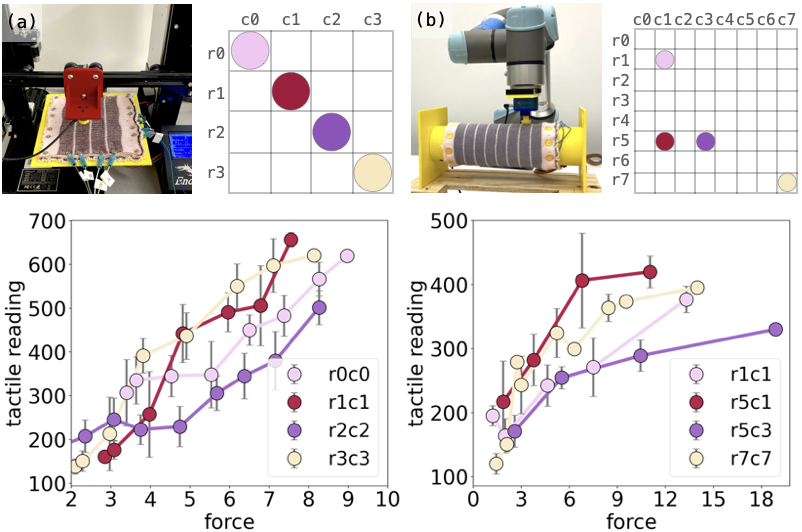}
    \vspace{-6mm}
    \caption{\Ztext{Tactile readings of our sensors from each taxel unit, \texttt{ricj} (located at i-th row and j-th column), are linearly correlated to the gradually increased applied normal forces at different locations on (a) flat and  (b) curved surfaces.}}
    \vspace{-6mm}
    \label{fig:flat-ft}
\end{figure}

The resistance at each taxel unit is decided by the contact pressure. 
We characterize the force-tactile transfer function on both a flat surface and a curved surface (Fig.~\ref{fig:flat-ft}).
We preset the indentation depth and static contact time, then for each taxel unit, we load, stay in static contact for 10 seconds, and unload the indenter. 
We gradually increase the indentation depth and record the quasi-static readings. 
\Ctext{}

\subsubsection{Flat surface}
As shown in setup image (Fig.~\ref{fig:flat-ft}a), a 4 $\times$ 4 tactile sensor is placed on a flat surface.
We show the characterization data at 4 locations in Fig.~\ref{fig:flat-ft}.
Tactile readings are approximately linearly-correlated
to the applied \Ztext{normal} forces at each taxel unit.
\Ctext{The correlation coefficients differ because of the varying number of margin resistors.}
\subsubsection{Curved surface}
We test the 8 $\times$ 8 tactile sensor placed on a curved surface, as shown in (Fig.~\ref{fig:flat-ft}b).
We show the characterization data at 4 locations in Fig.~\ref{fig:flat-ft}.
We also observe the approximate linear relationship
between the tactile readings and the applied \Ztext{normal} forces. 
However, the variance of the data is larger than the ones from the flat testbed. 
This is because, with a rigid indenter, the contact area degrades to smaller on a curved surface, such as point or line contact, which is less stable. 
But our sensor responds robustly to large-area contact such as human hands which will be shown in applications (Sec~\ref{sec:applications}).

\Ctext{
The minimum detectable force is around 2.5N, less than the average force of a finger tap~\cite{finger-force}.
We observed similar relative readings w.r.t. applied normal forces on different surfaces.
The experimental setup above can perform tactile-force calibration for precise force estimation.
}

\definecolor{myBurgundy}{rgb}{0.656,0.035,0.23}
\definecolor{myYellow}{rgb}{0.969,0.906,0.738}
\definecolor{myPink}{rgb}{0.953,0.777,0.957}
\definecolor{myPurple}{rgb}{0.582,0.313,0.746}
\definecolor{myBlue}{rgb}{0.617,0.691,0.805}

\begin{figure}[t]
    \centering
    \includegraphics[width=\linewidth]{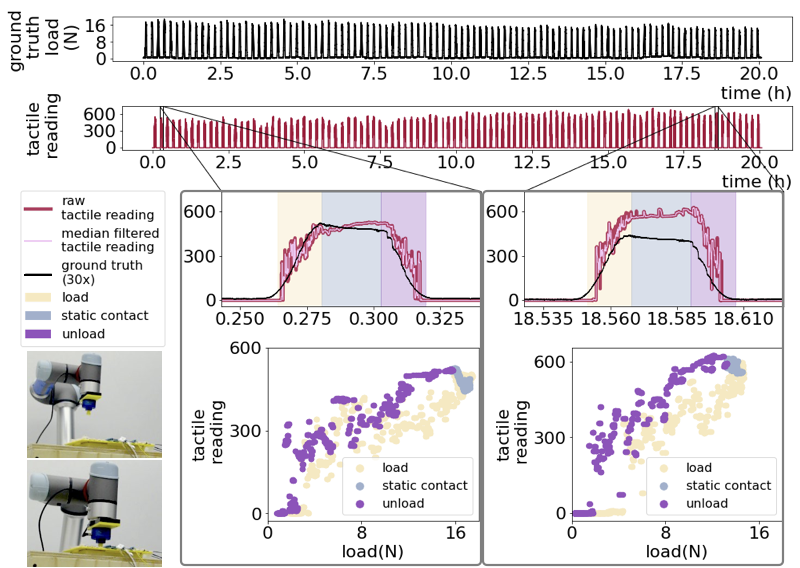}
    \caption[repeatability]{
    Repeatability test of our sensor. 
    The experimental setup is shown on the bottom left.
    Using force-torque sensor's z-axis readings as the ground truth load (black lines \tikz\draw[black,fill=black] (0,0) circle (0.8ex);), we repeat
    force loading and unloading on a single taxel unit and plot the corresponding tactile reading (burgundy lines \tikz\draw[black,fill=myBurgundy] (0,0) circle (0.8ex);). We zoom in two cycles of: 
    load \tikz\draw[black,fill=myYellow] (0,0) circle (0.8ex);),
    static contact \tikz{\draw[black,fill=myBlue] (0,0) circle (0.8ex);}, 
    and unload \tikz{\draw[black,fill=myPurple] (0,0) circle (0.8ex);} at the beginning and the end of the test respectively.
    \label{fig:repeatability}}
\end{figure}
\begin{figure}[t!]
    \centering
    \vspace*{-8mm}
    \includegraphics[width=\linewidth]{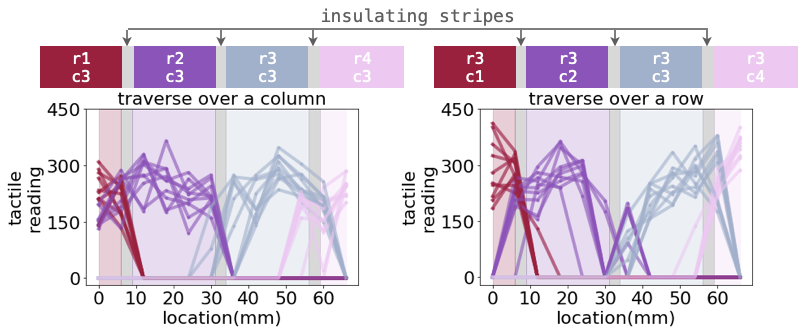}
    \caption{Contact localization test by traversing loads 10 times over a row or a column of our knitted sensor. Taxel units are alternatively activated over the row/column with \Ztext{approximate} Gaussian distributions. When the indenter loads on two adjacent taxel units simultaneously, two units are activated.}
    \label{fig:traverse}
    \vspace*{-5mm}
\end{figure}

\subsection{Repeatability}\label{sec:repeatability}
Long-lasting usage of sensors requires stable and repeatable sensor readings. 
To test \Ctext{the} robustness, we repeat the following cycle 100 times across 4 days (20, 30, 30, 20 cycles per day, respectively): \Ctext{6s} load at 0.5mm/s \Ctext{(3mm displacement)}, \Ctext{10s} static contact, \Ctext{6s} unload at 0.5mm/s, and 20s rest. The result is shown in Fig.~\ref{fig:repeatability}: our sensor responses in each cycle to the applied forces with repeatable behaviors. 
In the zoom-in views of two cycles from the beginning and end of the test, we show our sensor has approximately linear force-tactile behavior during loading and unloading, stable readings under stable contacts. \Ztext{We observe hysteresis between loading and unloading, and sensor reading drift caused by sensor's permanent deformation, which is not explicitly avoided but in an acceptable range for our applications.}

\subsection{Contact localization}\label{sec:localization}
A matrix-like sensor layout naturally embeds spatial information but the spatial resolution depends on the taxel unit sizes.
The insulating mesh layer in our sensor eliminates the cross-talk between adjacent taxels. 
Therefore, the contact at each taxel can be precisely located. 
We traverse and load forces over a row/column and show the tactile readings in Fig.~\ref{fig:traverse}.
The specifications of our 4 $\times$ 4 sensor is indicated in Table~\ref{table:sensor-specs} where the taxel unit size is 22mm $\times$ 22mm and the insulating margin is 3mm.
We set each step size as 6mm and load contact forces using a flat round indenter with 20mm diameter along a line. 
When the contact is within a taxel unit, only the corresponding tactile taxel responses. 
When the contact crosses two taxel units, both units respond. 
\Ztext{Tactile readings appear to be in approximate Gaussian distributions over locations from the plots.}

\begin{figure}[t!]
    \centering
    \includegraphics[width=\linewidth]{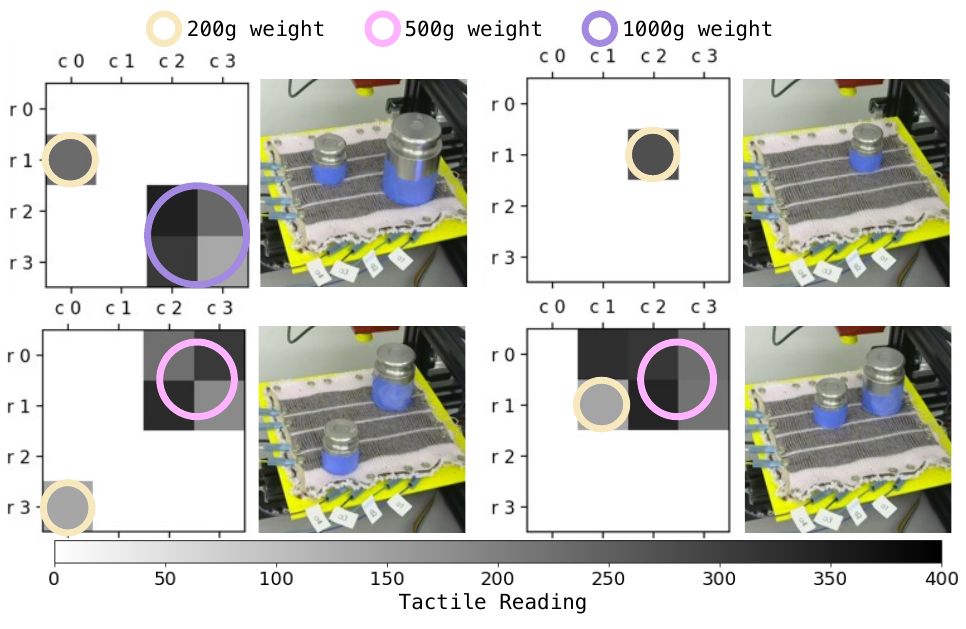}
    \vspace*{-7mm}
    \caption{Multi-contact detection \Ctext{as w}eights \Ctext{are placed} at different locations. \Ztext{Taxel units are colored corresponding to the tactile readings, as shown in the bottom colorbar}. Notice the ghosting at \texttt{r0c1} in the bottom right plot.}
    \label{fig:multi-contact}
\end{figure}

We also test measurement with multi-contact localization, as shown in Fig.~\ref{fig:multi-contact}. 
We place different weights at different locations and record sensor readings. 
The readings accurately reflect contact locations and contact forces. 
However we notice the ghosting in the bottom right plot at \texttt{r0c1}: a reading observed due to contacts at \texttt{r0c2}, \texttt{r1c1}, and \texttt{r1c2}.
Crosstalk, current leakages between channels, and ghosting are known issues in resistor matrices.
There are mechanisms to eliminate these issues.
Our experiments with the SPDT-mux approach~\cite{3DKnITS,knitui} show the reduction of ghosting is at the cost of accurate multi-contact readings.

\subsection{Behavior under different contact shapes}\label{sec:indenter}
\begin{figure}[t!]
    \centering
    \vspace*{-2mm}
    \includegraphics[width=\linewidth]{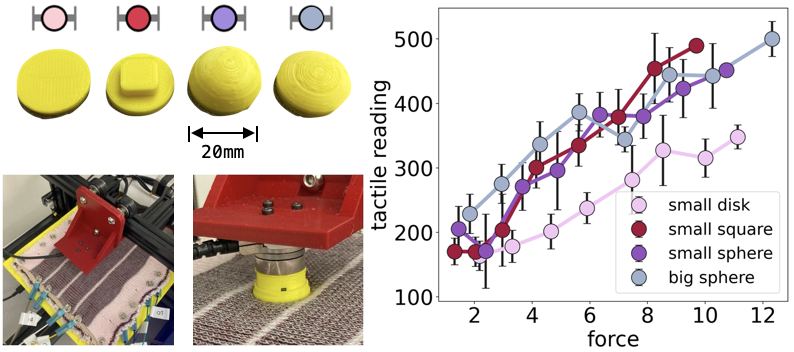}
    \vspace{-5mm}
    \caption{Quasi-static load test with different contact shapes. We test indenters with four shapes: a disk-shape, a small square, a small sphere, and a large sphere. The force-tactile behaviors are shown on the right.}
    \label{fig:diff-indenter}
    \vspace{-6mm}
\end{figure}

For all the characterization discussed above, we use a disk-shaped indenter (leftmost indenter in Fig.~\ref{fig:diff-indenter}). 
However, the size and shape of the contact also affect the deformation of the knitting layers and lead to different tactile readings. 
Therefore, we conduct the same force-loading experiment with four different indenters: a disk-shape, a small square, a small sphere, and a large sphere on the flat testbed. 
Shown in Fig.~\ref{fig:diff-indenter}, the last three indenters have similar readings but differ from the disk-shaped indenter, \Ctext{ which has a larger contact area}. 
\Ctext{Our sensor is sensitive to the distribution of the applied forces as it affects sensor deformations. The sub-taxel resolution can be used for future object recognition research.}

\subsection{Readout circuit performance}\label{sec:readouts-performance}
We use an Arduino microcontroller to control the multiplexers and read the analog signals from the sensors. 
To get tactile readings at each taxel unit,
we first write to the two 16-channel multiplexers with $3.4$ (Arduino \texttt{DigitalWrite()}time)$\times$4 (4 channels of multiplexer)$\times$2 (2 multiplexers)=27.2 $\mu$s delay, and then read with 400$\mu$s delay for stable reading. 
Decreasing the delay time will increase sample rates but increase crosstalk. Sample rates reported in Table~\ref{table:sensor-specs} had negligible crosstalk.
We measure the energy consumption of our readout system with a 5V power supply.
Our sensor's energy consumption (reported in Table~\ref{table:sensor-specs}) is low and comparable to when the microcontroller runs an idle program.






\section{Robot Applications}\label{sec:applications}
%

\begin{figure}[b!]
    \centering
    \vspace*{-2.5mm}
    \includegraphics[width=\linewidth]{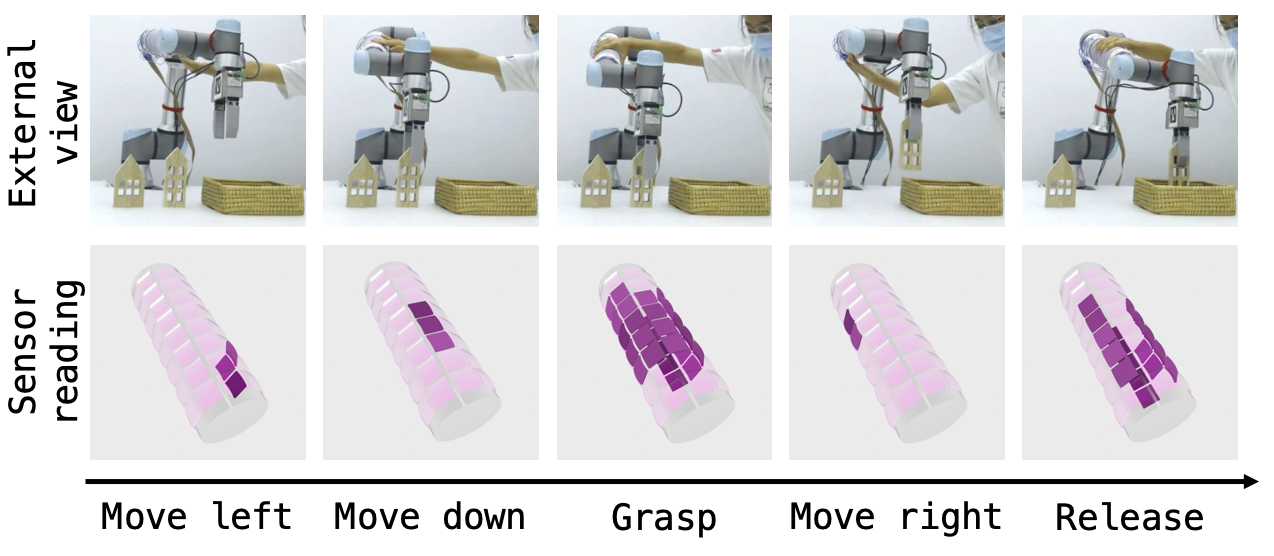}
    \vspace*{-6mm}
    \caption{\theSkin~on a UR5e robot arm. We use the tactile reading to control the robot's motion and the gripper's grasping to relocate an object. The robot arm moves in the opposite direction when detecting a ``push'' and the gripper will open/close when detecting a ``grab''.
    }
    \label{fig:app-robot-arm}
    \vspace{-4mm}
\end{figure}
\begin{figure}[b!]
    \centering
    \includegraphics[width=0.85\linewidth]{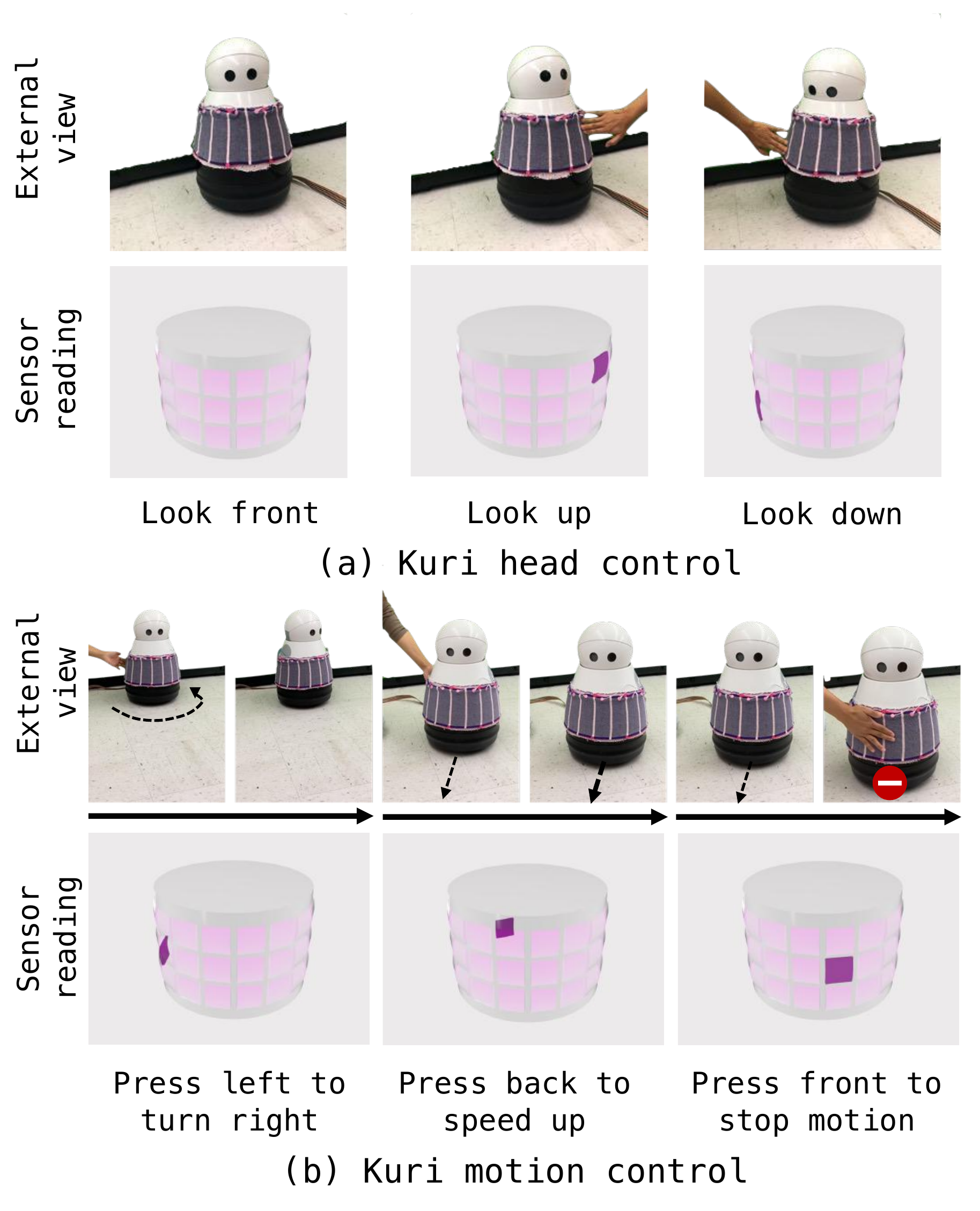}
    \vspace*{-3mm}
    \caption{\theSkin~on a Kuri robot for human-robot interaction. 
    (a): Kuri's head moves toward the direction that a human touches at.
    (b): we control Kuri's motion with touch. Kuri turns to the opposite direction to avoid contact, speeds up when being pushed from the back, and stops when detecting a contact in the front. 
    }
    \vspace*{-5mm}
    \label{fig:app-mobile-robot}
\end{figure}

Using \theSkin, we show closed-loop control for two human-robot interaction applications: a robot arm \Ctext{with human lead-through control} and a mobile robot interaction.

\subsection{\Ztext{Human lead-through control} with tactile feedback}\label{sec:ur5e}
We fabricate a cylinder-shaped tactile sensor, the 8$\times$8 prototype in Table~\ref{table:sensor-specs}, and instrument on the forearm link of a UR5e robot arm (Fig.~\ref{fig:teaser}). 
Human users \Ztext{can lead-through control the robot} to relocate an object (Fig~\ref{fig:app-robot-arm}).

Using the tactile reading, we implement (a) pin-point contacts controlling 3 DoF translational movements of the end-effector (EE), and (b) large-area gesture recognition controlling the gripper.
For (a), we directly control the velocity of EE given the contact locations and forces.
Velocity direction is from the contact point to the center of the cylinder. 
Velocity magnitude is linearly correlated to the detected force.
For (b), we control the gripper to open/close for grasping based on the current gripper state. As shown in Fig.~\ref{fig:app-robot-arm}, we can effectively control a robot to grasp an object, and deliver it to the target location with only tactile feedback from our sensor.

\subsection{Human-robot interaction with tactile sensing}\label{sec:kuri}
We fabricate a cone-shaped tactile sensor, the 3$\times$16 prototype in Table~\ref{table:sensor-specs}, and instrument on the body of Kuri, a mobile home robot (Fig.~\ref{fig:teaser}). 
We use \theSkin's tactile readings to implement two applications: head control and motion control (Fig.~\ref{fig:app-mobile-robot}). 

Kuri's ``eyes" have an embedded camera for live-streaming and capturing images.
We control the head motion of Kuri based on detected touch locations from \theSkin. 
We discretize the pitch motion as looking up, looking forward, and looking down from 3 rows, and the yaw motion into 16 different angles from 16 columns. 
When the touch is detected, the robot responds with the corresponding head motion to alter the camera view.

We can also control Kuri's motion based on tactile feedback.
Based on our definition, if contact happens at the front, the robot stops walking;
if contact happens at the side, the robot turns to the opposite direction to avoid collisions;
and if contact happens at the back, the robot speeds up assuming it is pushed by a human. 
We demonstrate the interaction with Kuri with our knitting sensor in Fig.~\ref{fig:app-mobile-robot} and show that Kuri successfully responds to human touch in an intuitive way.


\section{Conclusions}
In this work, we present \theSkin, a scalable, generalizable, and customizable machine-knitted tactile skin for robots. Our tactile skin leverages off-the-shelf functional yarns with machine-knitting techniques. It can sense contact locations and forces effectively. We characterize our tactile skin's force sensitivity, repeatability, contact localization, and readout circuit performance. Furthermore, we demonstrate the usage of our tactile skins for lead-through control of a robot arm and interaction with a mobile robot.

For future work, we would like to further apply our tactile skins on robots with various shapes and sizes~\cite{autoknit} for safer human-robot interactions. Taxel unit placement can be designed for non-uniform dimension and distribution~\cite{visualknit}. We also plan to improve the circuit design and add post-processing~\cite{crosstalk-postprocess} to solve the multi-contact ghosting effect. 


\addtolength{\textheight}{-3cm}   






\bibliographystyle{bib/IEEEtran}
\bibliography{bib/IEEEabrv, bib/ref, bib/more-ref}

\end{document}